\documentclass[a4paper,twoside]{article}

\usepackage{epsfig}
\usepackage{subcaption}
\usepackage{calc}
\usepackage{amssymb}
\usepackage{amstext}
\usepackage{amsmath}
\usepackage{amsthm}
\usepackage{multicol}
\usepackage{pslatex}
\usepackage{natbib}
\usepackage{todonotes}
\usepackage{setspace}
\usepackage{hyperref}
\usepackage{SCITEPRESS}     


\begin{document}

\title{SemSegDepth: A Combined Model for Semantic Segmentation and Depth Completion}

\author{\authorname{Juan Pablo Lagos\sup{1}, Esa Rahtu\sup{1}\orcidAuthor{0000-0001-8767-0864}, }
\affiliation{\sup{1}Tampere University, Tampere, Finland}
\email{\{juanpablo.lagosbenitez, esa.rahtu \}@tuni.fi}
}


\abstract{Holistic scene understanding is pivotal for the performance of autonomous machines. In this paper we propose a new end-to-end model for performing semantic segmentation and depth completion jointly. The vast majority of recent approaches have developed semantic segmentation and depth completion as independent tasks. Our approach relies on RGB and sparse depth as inputs to our model and produces a dense depth map and the corresponding semantic segmentation image. It consists of a feature extractor, a depth completion branch, a semantic segmentation branch and a joint branch which further processes semantic and depth information  altogether. The experiments done on Virtual KITTI 2 dataset, demonstrate and provide further evidence, that combining both tasks, semantic segmentation and depth completion, in a multi-task network can effectively improve the performance of each task. Code is available at \url{https://github.com/juanb09111/semantic_depth}.}

\onecolumn \maketitle \normalsize \setcounter{footnote}{0} \vfill
\global\csname @topnum\endcsname 0
\global\csname @botnum\endcsname 0

\section{\uppercase{Introduction}}
\label{sec:introduction}
Computer vision and holistic scene understanding have become pivotal topics as we intend to provide machines with autonomous capabilities. When we, as humans, see things we unconsciously assign multiple attributes to what we see and we also perform multiple tasks simultaneously. For instance, we can effectively assess the distance of the objects we see, the quantity, the size, the texture, etc. all at once. We are also capable of understanding the world around us in its semantic complexity.  On the other hand, machines can outperform humans in several tasks individually. That is the case for tasks such as object detection \citep{ren2016faster}, \citep{zhai2017feature}, \citep{redmon2018yolov3}, semantic segmentation \citep{unet}, \citep{deeplab_Ex}, \citep{lin2016refinenet} and/or depth estimation \citep{chen2020learning}, \citep{godard2019digging}, \citep{guizilini20203d}, where machines have been able to successfully carry out those tasks individually. However, when it comes to performing multiple tasks, machines are still lagging behind, in comparison to humans.
\begin{figure}
    \centering
    \includegraphics[width=0.48\textwidth]{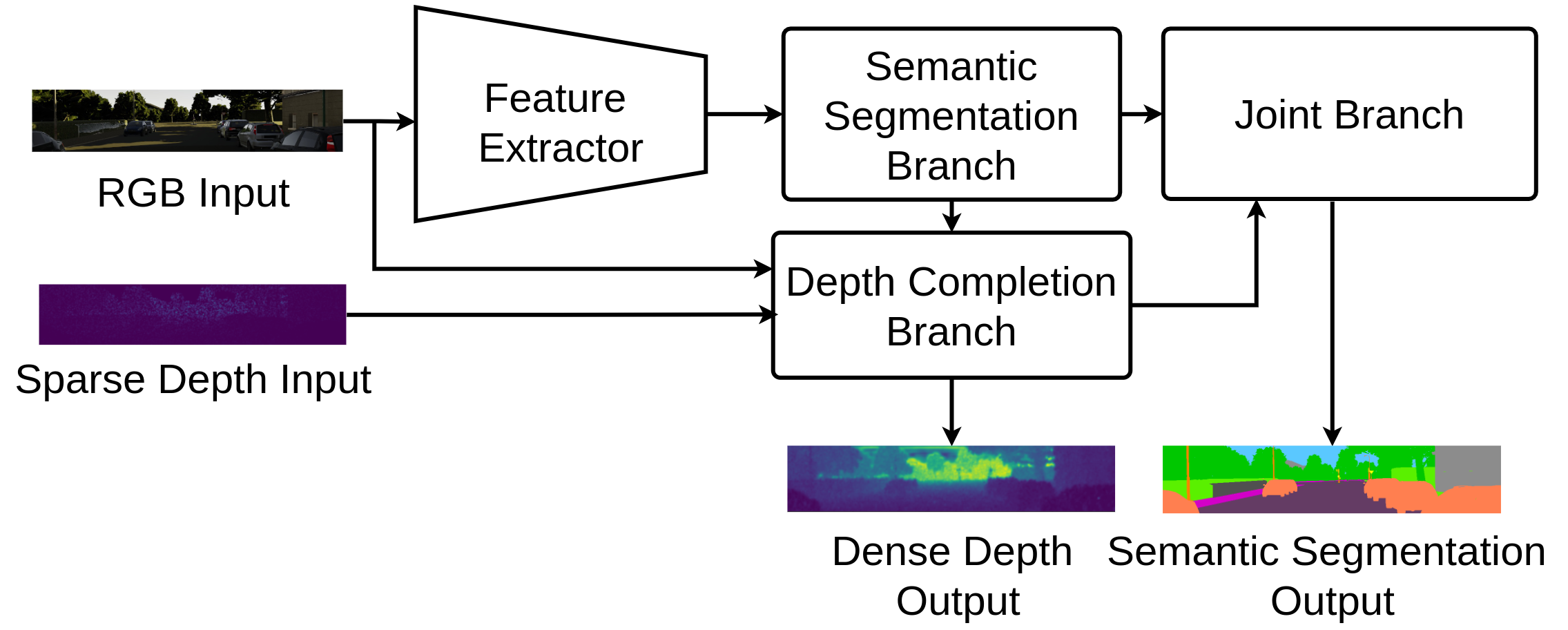}
    \caption{Overview of our proposed SemSegDepth architecture. Our model produces a dense depth map and semantics prediction given an RGB image and sparse depth as input.}
    \label{fig:overview}
\end{figure}

In an attempt to provide a more holistic approach to the problem of scene understanding, multi-task networks have become a highly active field of research in computer vision. In addition to provide a more complete representation of a scene, there is growing evidence that multi-task networks can improve the performance of each individual task \citep{liebel2018auxiliary}. Panoptic segmentation, for example, combines instance segmentation, object detection and semantic segmentation \citep{Effps}, \citep{cheng2020panopticdeeplab}, \citep{wang2020pixel}, \citep{weber2020singleshot}. To our best knowledge, only few methods have combined semantic segmentation and depth completion \citep{8462433}, \citep{Simultaneous}. As compared to other applications, e.g. panoptic segmentation, combining semantic segmentation and depth completion poses additional challenges such as processing heterogeneous data jointly, since semantic segmentation relies on RGB images while depth completion relies on sparse depth data.

Semantic segmentation refers to the task of assigning a semantic label to every single pixel in an image, e.g. determining whether a pixel in an image belongs to a "car", "person", "bike" or "background". On the other hand, depth estimation, more specifically depth completion, predicts the distance of every pixel in an image, where, in most cases, a sparse depth input is provided. In applications such as autonomous driving, combining semantic segmentation and depth completion can improve the performance of the system as a whole significantly, as the machines would not only understand their surroundings semantically, but also, they would have knowledge about the proximity of the things on a given scenario.

In this paper we proposed a new end-to-end multi-task network for performing semantic segmentation and depth completion jointly. We combine two bench-marking models, namely, we use a modified version of the depth completion network proposed by \citet*{chen2020learning} as well as a modified version of EfficientPS \citep{Effps}. An overview of our model SemSegDepth is shown in Figure \ref{fig:overview}. It consists of a feature extractor, a semantic segmentation branch, a depth completion branch and a joint branch. The feature extractor is a resnet50 network \citep{he2015deep}  wrapped in a Feature Pyramid Network (FPN) \citep{lin2017feature}. Our semantic segmentation branch is based on the semantic segmentation branch of the EfficientPS architecture. The depth completion branch extends \citep{chen2020learning} by adding semantic logits as input, and finally, the joint branch further processes semantic and depth information altogether. We trained and evaluated our model on Virtual KITTI 2 \citep{cabon2020virtual} and demonstrated that our SemSegDepth model improves the performance for both tasks, semantic segmentation and depth completion.

\section{\uppercase{Related Works}}

\subsection{Semantic Segmentation}
Semantic segmentation takes image classification task to a pixel level. Fully convolutional networks have previously been used  to perform dense predictions for pixel-wise segmentation \citep{DBLP}. During the last decade encoder-decoder architectures such as UNet \citep{unet} became popular and achieved the state of art using what today can be considered rather simple architectures based deep convolutional networks (DCNN) that were capable of restoring the original spacial resolution with a series of upsampling layers in an end-to-end manner. However, traditional upsampling layers such as bi-linear upsampling or deconvolutional layers are computationally expensive. 

Atrous convolution is a more efficient alternative and architectures such as DeepLab \citep{deeplab} pioneered using atrous convolution in the context of pixel-wise semantic segmentation using DCNNs. DeepLab also introduced the concept of "atrous spatial pyramid pooling" (ASPP) to enhance the network's capability of representing objects of different sizes. Later in \citep{deeplab_Ex} ASPP would be optimized by using depth-wise separable convolution which would result in a faster yet stronger network. Alternatively, \citet*{R2} proposed a novel and different approach to deal with the problem of scale variation in images, by using reconstructed depth from stereo images and a pixel-wise scale selection multiplexer which provides a scale-invariant image representation successfully used by a classification sub-network that finally outputs the semantic segmentation map. Other architectures \citep{effnet}, \citep{xception}, \citep{Effps} would also benefit from depth-wise separable convolutional layers which are many times faster than traditional convolutional layers. 

Another approach adopted in convolutional neural networks (CNN) is called gated convolutions which are based on linearizing belief nets (LBNs)  \citep{LBN} that are capable of modeling a deep neural network (DNN) as linear units that can be turned on and off in a non-deterministic fashion reducing the vanishing gradient problem. LBNs were later used for language modeling by \citet*{lmodeling} and further applied in the context of semantic segmentation by \citet*{SCNN} whose work tackles the problem with two branches, one of which processes the shape while the other branch processes semantic information in a classical way, and the two branches are connected with gating mechanisms.

\subsection{Depth Completion}
Neural networks have been largely used to produce dense depth maps out of sparse data provided by depth sensors such as Lidar and the vast majority of those networks also use RGB images for guidance \citep{imran2019depth}, \citep{DDP} \citep{xu2019depth},  \citep{huang2020hmsnet},  \citep{tang2019learning}.

The lack of ground truth dense depth maps poses a challenge for supervised learning approaches. Existing datasets like Virtual KITTI 2 \citep{cabon2020virtual} provide synthetic data including dense depth maps ground truth. However, that is not the case in realistic scenarios where only sparse ground truth is available. Other approaches like \citep{ma2018selfsupervised} are capable of learning a mapping from sparse depth and images to dense depth with no need of dense depth maps as ground truth. It is also possible to learn depth features by using surface normals as in \citep{DeepLiDAR} and \citep{Deep}.

One of the challenges of working with 3D data is its non-grid nature and therefore traditional CNNs simply do not work unless the 3D points are mapped on to a 2D space. Motivated by this problem \citet*{Wang_2018} introduced what they called Parametric Continuous Convolution to learn features over non-grid data.

Making use of the recent continuous convolution proposed in \citep{Wang_2018}, \citet*{chen2020learning} introduced a neural network block which extracts 2D and 3D features jointly. Such block consists of two branches running in parallel. One of the branches processes RGB features while the other branch uses continuous convolution over 3D points and finally the outputs of both branches are fused together. By stacking the same block N times they managed to effectively produce a dense depth outperforming the state of the art in 2020. Our proposed model builds upon the architecture proposed by \citet*{chen2020learning} for performing depth completion as described in section \ref{sec:dc}. 

\subsection{Multi Task Leaning}
A CNN can effectively be trained to produce multiple outputs corresponding to different tasks. In the context of image processing, tasks such as object detection and semantic segmentation have been tackled successfully \citep{6247739}, \citep{Effps}, \citep{kim2020video}. Depth estimation and semantic segmentation have also been combined in one CNN as in \citep{EigenF14}, \citep{hazirbas16fusenet}, \citep{KendallGC17}, \citep{Simultaneous}

Multi task networks have shown to achieve better results as whole in terms of their capability to provide a more holistic representation, but also the performance of each one of the tasks improves as a result of having a multi task CNN. Inspired by this approach, \citet*{liebel2018auxiliary} introduced the concept of "auxiliary tasks", they are side tasks that are less relevant for a given application but that potentially improved the performance of the core tasks. Moreover, \citet*{R3} suggested that certain visual tasks contain underlying common and supplementary features, meaning that high level representations of an input for a specific task, may contain relevant information for solving a different task, as long as the tasks are related to one another. 

More recently, \citet*{R1} proposed a multi task network for semantic segmentation and depth completion which exploits the geometric relationship between the two tasks by introducing the concept of semantic objectness, used as a constrain that describes the correlation between the semantic and the actual depth.

\section{\uppercase{Architecture}}

\subsection{Overview}

\begin{figure*}[h]
    \centering
    \includegraphics[width=1\textwidth]{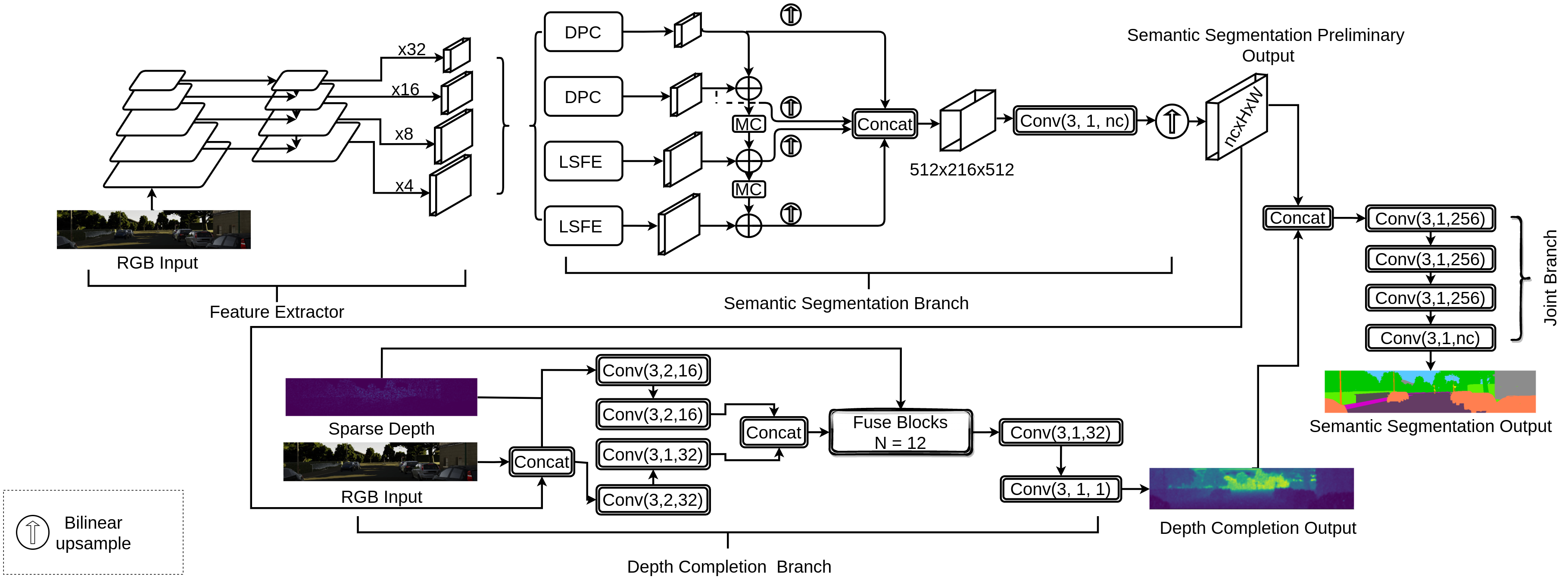}
    \caption{Diagram of the SemSegDepth architecture. The convolutional layers shown in  this diagram follow the notation Conv($k$,$s$,$c$) where $k$ refers to a $k \times k$ convolutional kernel, $s$ is the stride, and $c$ is the number of output feature channels.}
    \label{fig:semseg_depth}
\end{figure*}

We propose a CNN which takes as inputs a single RGB image and a sparse depth image, and returns the corresponding semantic segmentation image and a dense depth map in an end-to-end manner. The complete diagram of our model is shown in Figure \ref{fig:semseg_depth}. Our model consists of one feature extractor backbone, two task-specific branches, one of which is designed for performing semantic segmentation and another one which performs depth completion, and one joint branch which combines semantic and depth information. The two task-specific branches are  intercommunicated at specific points. The semantic segmentation branch is based on a neural network known as "EfficientPS" for panoptic segmentation \citep{Effps}, from where we neglect the instance segmentation branch, while the depth completion branch is based on a fusion network introduced by \citep{chen2020learning} which extracts joint 2D and 3D features. Thus, our proposed architecture  combines  two bench-marking  models to perform semantic segmentation and depth completion jointly.

\subsection{Backbone}

The backbone, as shown in Figure \ref{fig:semseg_depth}, is a resnet50 feature extractor \citep{he2015deep} wrapped in a FPN \citep{lin2017feature} for extracting intermediate features from the backbone in order to have feature maps at multiple scales. More specifically, the FPN returns four feature maps that are down-sampled, with respect to the input, by a factor of $\times 4$, $\times 8$, $\times 16$ and $\times 32$. These features are then fed to the semantic segmentation branch.

\subsection{Semantic Segmentation Branch}

The semantic segmentation head follows the architecture of the semantic segmentation branch proposed by \citet*{Effps}. The inputs of this branch are the four different outputs of the feature extractor, that is, four feature maps, each one with a different spacial resolution. This semantic segmentation head aims at capturing large-scale features as well as small-scale features and then on a later stage, such feature maps at different scales are aggregated. This branch returns semantic logits as output and the resolution of the output is  $nc \times H \times W$ where $nc$ corresponds to the number of classes.

In order to extract large-scale features, we use a Large Scale Feature Extractor (LSFE) module which consists of a stack of three layers of $3 \times 3$ convolutions and produces a feature map with $128$ filters. For extracting small-scale features, we used what is known as Dense Prediction Cells (DPC) \citep{chen2018searching} which is a modified version of ASPP \citep{deeplab}.

Finally, in order to reduce the mismatch between small-scale features and large-scale features, we used a Mismatch Correction Module (MC). It consists of a stack of three layers of $3 \times 3$ convolutions and one bilinear upsampling layer at the very end.

\subsection{Depth Completion Branch}\label{sec:dc}

Our depth completion branch is a modified version of the depth completion network proposed by \citet*{chen2020learning}. To begin with, our depth completion branch receives as input, not only sparse depth and RGB as in \citep{chen2020learning}, but it has one more input which corresponds to the preliminary output of the semantic segmentation branch. Thus, we are embedding semantic information into our depth completion branch.

The sparse depth, RGB and the semantic segmentation preliminary output are concatenated and passed through a stack of two $3 \times 3$ convolutional layers. The sparse depth image alone is also passed through a stack of two $3 \times 3$ convolutional layers. Then, the two outputs are concatenated and the resulting tensor, as well as the sparse depth ground truth, are the inputs of a stack of $N$ $2D-3D$ Fuse Blocks \citep{chen2020learning}. Finally, the resulting output of the Fuse Blocks passes through two convolutional layers for further refinement. The output of this branch is a fully dense depth map, that is, an image where every pixel is assigned a value of depth.

\subsection{Joint Branch}

Finally, in order to use depth information as guidance for the semantic segmentation branch, we concatenate the output of the depth completion branch and the preliminary output of the semantic segmentation branch. The result is passed through a joint branch that processes semantic and depth information altogether. The purpose of the joint branch is to further process the semantic segmentation preliminary output guided by depth information. It consists of a stack of four $3 \times 3$ convolutional layers. The output of the joint branch is the corresponding $nc \times H \times W$ semantic logits based on which we calculate the loss as described in section \ref{semseg_sec}.
\subsection{Loss Functions}

\paragraph{Semantic Segmentation}\label{semseg_sec}

For semantic segmentation we computed the cross-entropy loss for every pixel. The loss for a pixel $i$ is defined as:

\begin{equation}\label{eq1}
    L_{semantic} = -\sum_{i} p_i \log \hat{p_i},
\end{equation}

where $i$ is the pixel index, $p_i$ is the ground truth and $\hat{p_i}$ is the log Softmax value of the predicted probability for pixel $i$. The log Softmax function is defined as:

\begin{equation}\label{eq2}
    LogSoftmax(x_i) = \log \left(\frac{exp(x_i)}{\sum_{j} exp(x_j)}\right)
\end{equation}

\paragraph{Depth Completion}

For depth completion we used the squared error average across all the pixels in the image for which the ground truth labels were available. The loss function for depth completion is then defined as:

\begin{equation}\label{eq3}
    L_{depth} = \frac{1}{N}\sum_{i}(\hat{y_i} - y_i)^2,
\end{equation}

where $N$ is the number of pixels, $\hat{y_i}$ and $y_i$ are the predicted value and the ground truth for pixel $i$, respectively.

\paragraph{Joint Loss} In addition using a loss per task, we implemented a joint loss function in order to leverage the correlation between both tasks, namely semantic segmentation and depth completion. The joint loss is simply the sum of each individual loss as follows:

\begin{equation}\label{eq4}
    L_{joint} =  L_{semantic} + L_{depth}.
\end{equation}

\section{Experiment Setup}

\subsection{Implementation Details}
We implemented the network on PyTorch  and used PyTorch \textit{DistributedDataParallel} for data parallelism during training. We used one machine with four 16GB graphics processing units (GPU) for training. We optimized the loss function using the stochastic gradient descent (SGD) with an initial learning rate set to $16 \times 10^{-4}$, momentum set to $0.9$ and weight decay set to $5 \times 10^{-5}$.

\subsection{Dataset} The experiments were done on the Virtual KITTI 2 dataset \citep{cabon2020virtual}. Virtual KITTI 2 is a synthetic video dataset which provides ground truth annotations for multiple tasks, namely, instance segmentation, semantic segmentation, multiple object tracking (MOT), optical flow, depth estimation, object detection and camera pose. It also provides stereo images for every scene. Besides, every sequence is recreated with subtle changes in the viewing angles, more specifically, $\pm 15^{\circ} and \pm 30^{\circ}$ horizontal rotations and changes in the weather conditions such as foggy, cloudy, rainy, morning and sunset. We used $500$ samples for training, $125$ for evaluation, and $200$ samples for testing.

Virtual KITTI 2 provides fully dense depth maps as depth ground truth, meaning that the depth values are provided for every single pixel in the input image. However, in order to reproduce real conditions where the ground truth depth is acquired with sensors such as LIDAR, which can only provide sparse values within a given range, we filtered out all the points exceeding a distance of $50$ meters and then we randomly sampled the ground truth. Hence, our synthetic depth ground truth consists of sparse depth images containing $8000$ depth values per image, where each point is within a range of $50$ meters. We cropped all the images to a resolution of $200 \times 1000$ ($H \times W$).

\subsection{Evaluation}

\begin{figure*}
\centering
\begin{tabular}{ccc}
{\includegraphics[width = 1.9in]{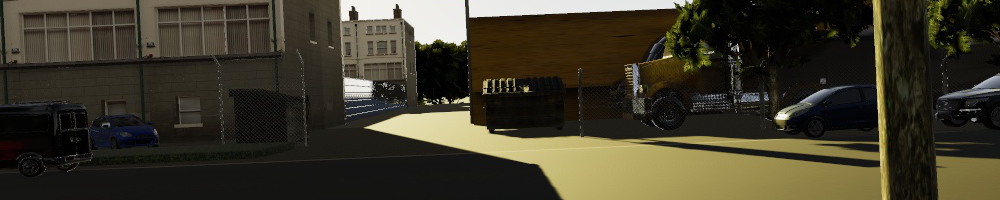}}&
{\includegraphics[width = 1.9in]{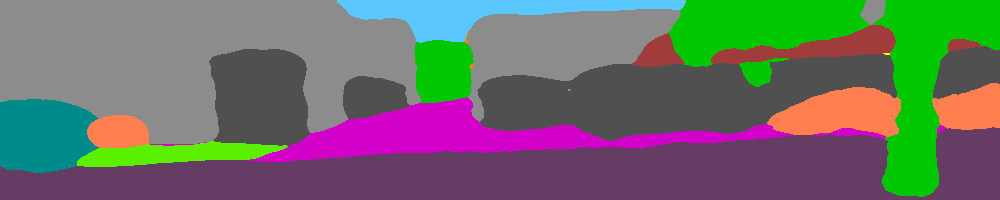}} &
{\includegraphics[width = 1.9in]{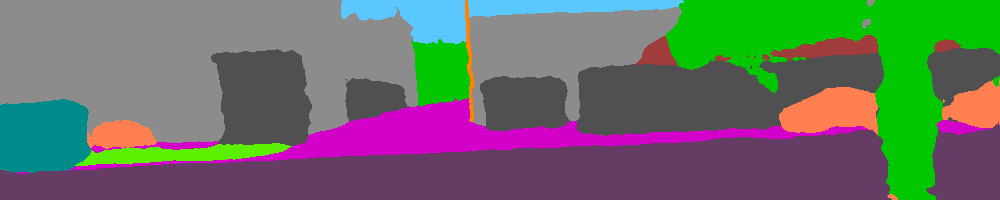}}\\

{\includegraphics[width = 1.9in]{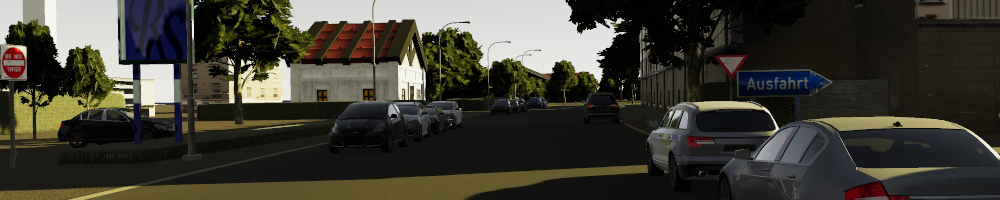}}&
{\includegraphics[width = 1.9in]{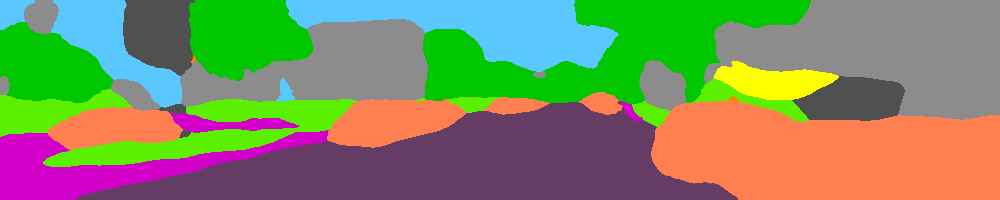}} &
{\includegraphics[width = 1.9in]{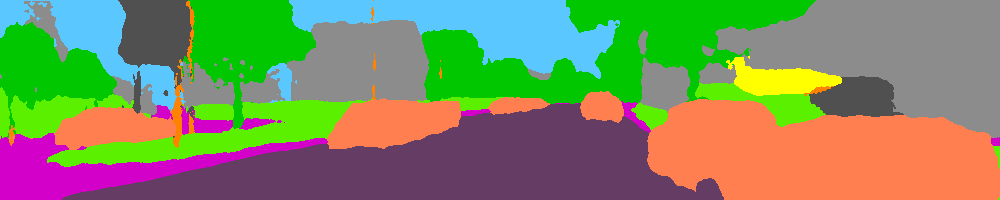}}\\

\subcaptionbox{RGB\label{ca1}}{\includegraphics[width = 1.9in]{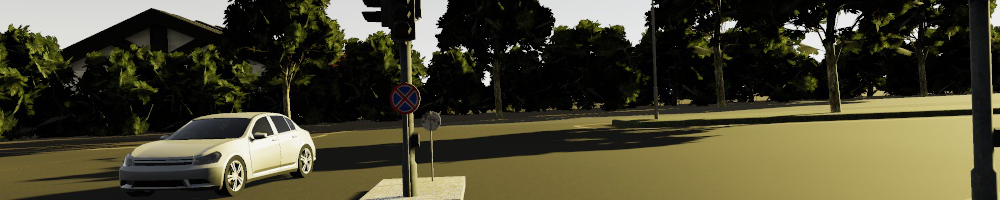}}&
\subcaptionbox{SemSegNet\_b\label{b1}}{\includegraphics[width = 1.9in]{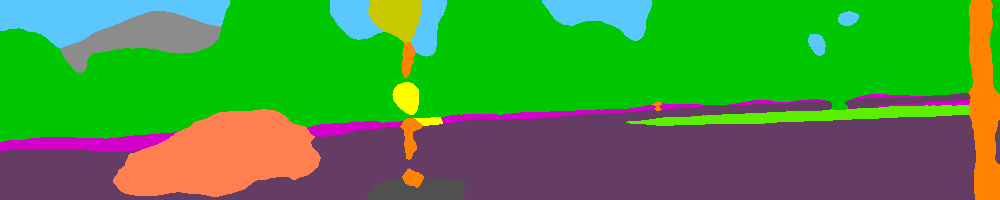}} &
\subcaptionbox{SemSegDepth (ours)\label{c1}}{\includegraphics[width = 1.9in]{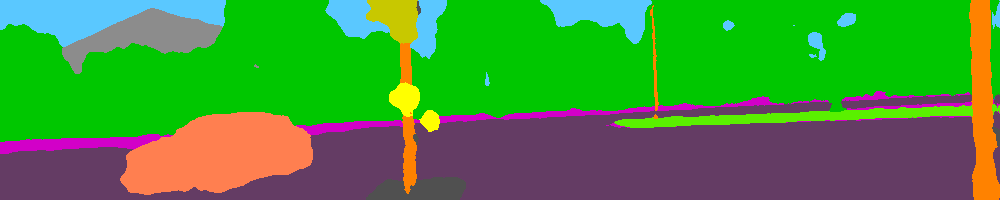}}\\
\end{tabular}
\caption{Semantic Segmentation results on Virtual KITTI 2. Column \textit{a} shows the RGB image input, column \textit{b} shows the semantic segmentation results using the baseline model SemSegNet\_b and column \textit{c} shows the the semantic segmentation  results using our model SemSegDepth.}
\label{semseg_res}
\end{figure*}

\begin{figure*}
\centering
\begin{tabular}{ccc}
{\includegraphics[width = 1.9in]{imgs/semseg_depth/00170_rgb.png}}&
{\includegraphics[width = 1.9in]{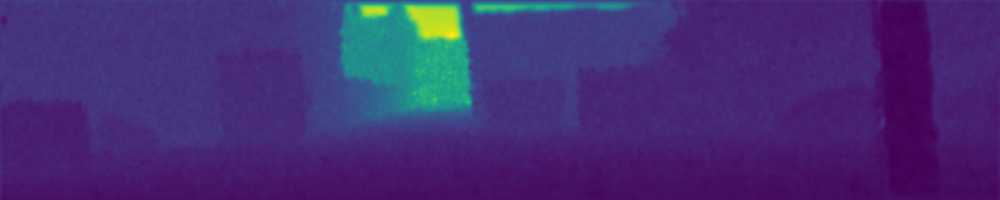}} &
{\includegraphics[width = 1.9in]{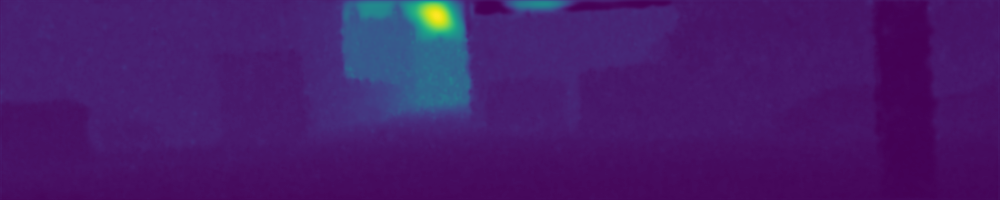}}\\

{\includegraphics[width = 1.9in]{imgs/semseg_depth/00256_rgb.png}}&
{\includegraphics[width = 1.9in]{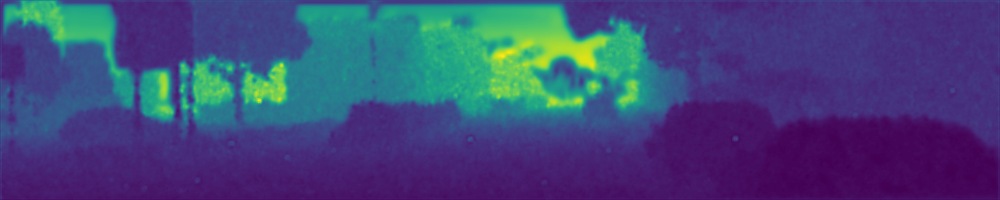}} &
{\includegraphics[width = 1.9in]{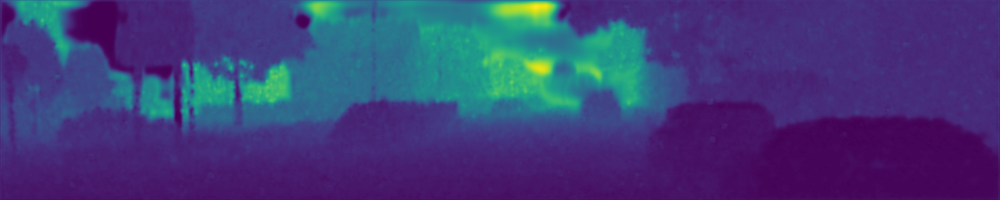}}\\

\subcaptionbox{RGB\label{ca2}}{\includegraphics[width = 1.9in]{imgs/semseg_depth/00411_rgb.png}}&
\subcaptionbox{DepthNet\_b\label{b2}}{\includegraphics[width = 1.9in]{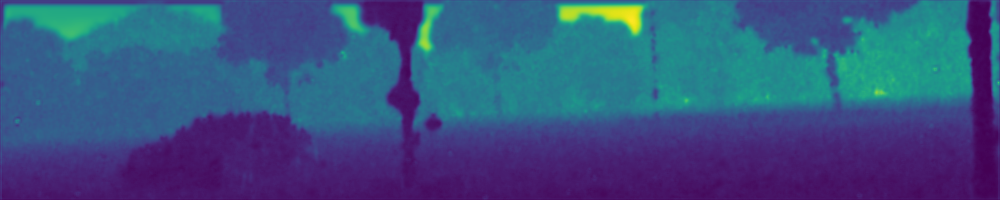}} &
\subcaptionbox{SemSegDepth (ours)\label{c2}}{\includegraphics[width = 1.9in]{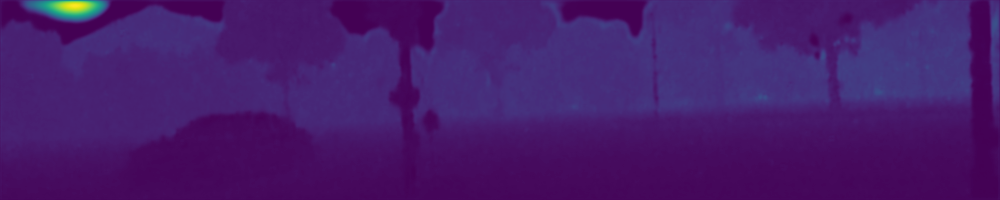}}\\
\end{tabular}
\caption{Depth completion results on Virtual KITTI 2. Column \textit{a} shows the RGB image input, column \textit{b} shows the depth completion results using the baseline model DepthNet\_b and column \textit{c} shows the the depth completion results using our model SemSegDepth.}
\label{depth_res}
\end{figure*}

In order to be able to evaluate the performance of our model, we used the mean Intersection-over-Union (mIoU) for evaluating the performance of semantic segmentation and root mean squared error (RMSE) for depth completion. The mIoU is defined as:

\begin{equation}\label{eq5}
    mIoU =  \frac{1}{nc}\sum_l \frac{TP_l}{TP_l+FN_l+FP_l},
\end{equation}

where $nc$ is the number of classes, $TP_l$, $FN_l$ and $FP_l$ are the number of true positive, false negative and false positive pixels respectively, labeled as class $l$. The RMSE metric is defined as:

\begin{equation}\label{eq6}
    RMSE = \sqrt{(1/N)\sum_{i}(\hat{y_i} - y_i)^2},
\end{equation}

 where $N$ is the number of pixels, $\hat{y_i}$ and $y_i$ are the predicted value and the ground truth for pixel $i$, respectively.

\subsection{Baseline}

In our work, the main purpose is to quantify the performance improvement of each individual task by having them as joint tasks in one single model. Therefore, our baseline consists of two models, one for each task. In this section, we describe each baseline model.

\paragraph{SemSegNet\_b} This is our semantic segmentation baseline. As proposed by \citet*{Effps}, the semantic segmentation task is carried out by an architecture which is composed by a feature extractor and a semantic segmentation  branch as shown in Figure \ref{fig:semseg_depth}. The depth completion branch and the joint branch are removed from this model.

\paragraph{DepthNet\_b} This baseline network is the model proposed by \citet*{chen2020learning} which corresponds to the depth completion branch of the model shown in Figure \ref{fig:semseg_depth} without concatenating the "Semantic Segmentation Preliminary Output" at the input. Hence, the only inputs are the RGB Input and the Sparse Depth.

\paragraph{}We compared our model to this baseline networks and the results are presented in section \ref{results}. We also present the results obtained with different configurations of our model as ablation studies in section \ref{ablation}.

\subsection{Results} \label{results}

The quantitative results for semantic segmentation and depth completion are shown in Table \ref{tab:semseg}. The best performance is highlighted in bold letters.

\begin{table}[h]
\caption{Results of our model compared to baseline networks.}\label{tab:semseg} \centering
\begin{tabular}{|c|c|c|}
  \hline
  Method & mIoU  & RMSE(mm) \\
  \hline
  SemSegNet\_b  & 0.520 & - \\
  \hline
  DepthNet\_b  & - & 580.2 \\
  \hline
  \textbf{SemSegDepth (ours)}  & \textbf{0.5932} & \textbf{458.2} \\
  \hline
\end{tabular}
\end{table}

As shown in Table \ref{tab:semseg}, our model outperforms each one of the baseline networks in every specific task. By combining both tasks, depth completion and semantic segmentation, our model achieved a significant improvement in the mIoU metric as compared to the semantic segmentation baseline model SemSegNet\_b. On the other hand, there was also a major improvement in the depth completion task. Qualitatively, the results are shown in Figure \ref{semseg_res} for the semantic segmentation task and in Figure \ref{depth_res} for the depth completion task.

\subsection{Ablation Studies} \label{ablation}

In addition to the baseline models, namely SemSegNet\_b and DepthNet\_b, we also designed four other networks which correspond to slight modifications of our model. These networks can be understood as intermediate steps from the baseline networks to our final model. In this section we describe each one of these models.

\begin{figure}[h]
    \centering
    \includegraphics[width=1\textwidth]{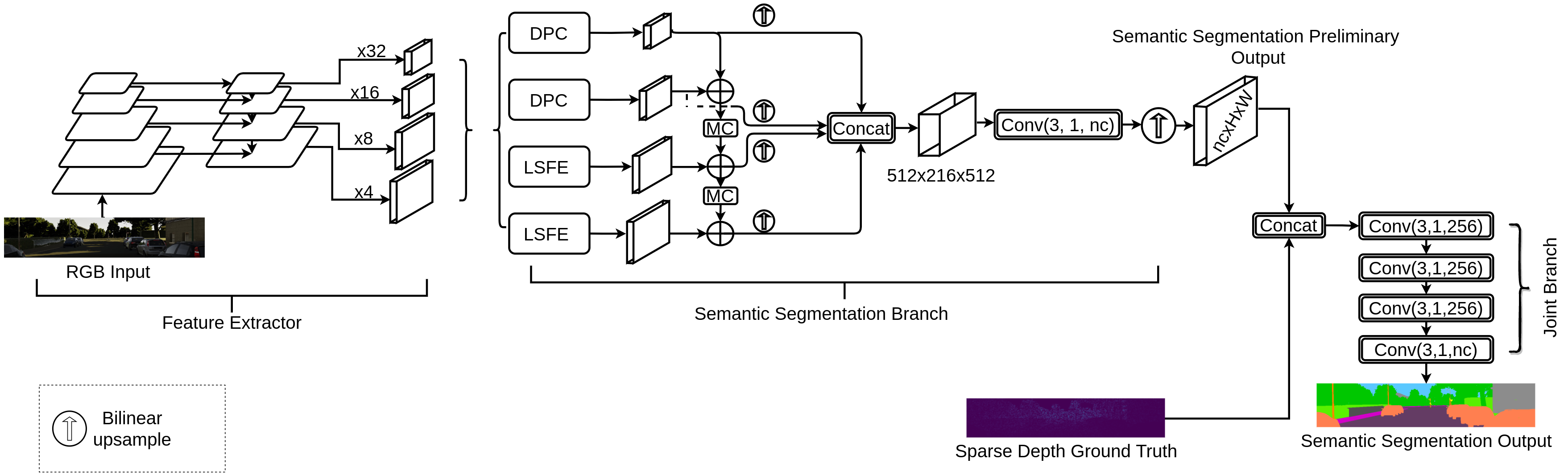}
    \caption{SemNet\_depth\_gt architecture.}
    \label{fig:app1}
    \end{figure}

\paragraph{SemNet\_depth\_gt} This is an extension of the semantic segmentation baseline network  SemSegNet\_b. This model  is based on the model we proposed, shown in Figure \ref{fig:semseg_depth}, and modified by removing the depth completion branch. The input to the joint branch is the concatenation of the Semantic Segmentation Preliminary Output and the depth completion sparse depth ground truth. The purpose behind this model is to evaluate whether or not providing depth information to the semantic segmentation baseline network can improve the performance for the task of semantic segmentation. The architecture of this model is shown in Figure \ref{fig:app1}.

\begin{figure}[h]
    \centering
    \includegraphics[width=1\textwidth]{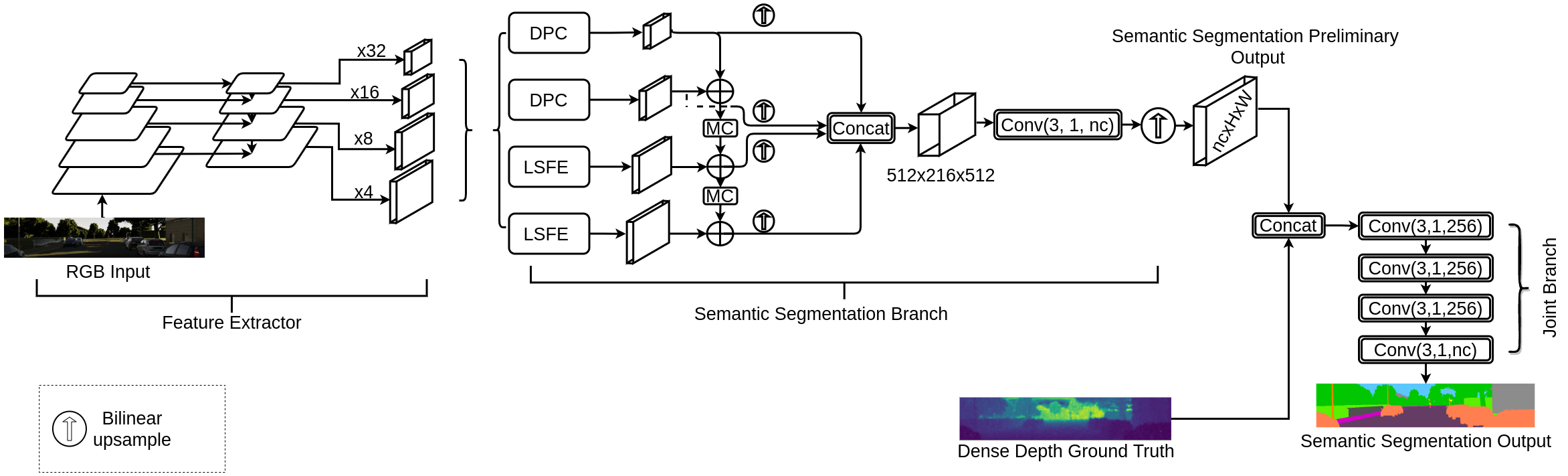}
    \caption{SemNet\_depth\_dense\_gt architecture.}
    \label{fig:app6}
    \end{figure}

\paragraph{SemNet\_depth\_dense\_gt} In contrast to SemNet\_depth\_gt, in this model, the input to the joint branch is the concatenation of the Semantic Segmentation Preliminary Output and a dense depth map ground truth. Figure \ref{fig:app6} shows the architecture of this model.

\begin{figure}[h]
    \centering
    \includegraphics[width=1\textwidth]{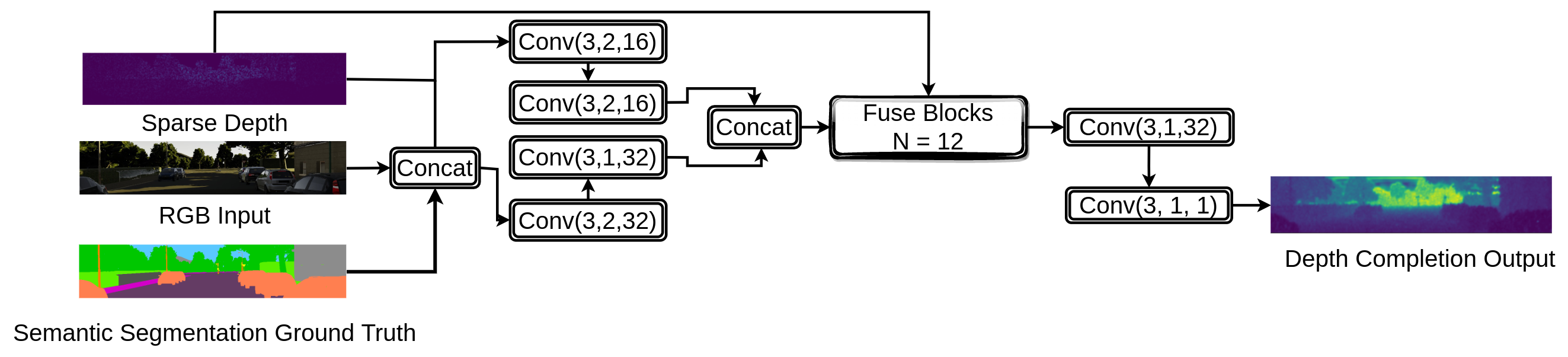}
    \caption{DepthNet\_semantic\_gt architecture.}
    \label{fig:app2}
    \end{figure}

\paragraph{DepthNet\_semantic\_gt} As shown in Figure \ref{fig:app2},  this is a modification of the depth completion baseline network DepthNet\_b. Similar to SemNet\_depth\_gt, we wanted to study whether or not providing reliable semantic information could improve the performance of the depth completion task alone. Therefore, we added one more input to the DepthNet\_b network, it corresponds to the semantic segmentation ground truth image which is concatenated with the RGB input at the first concatenation layer. 

\begin{figure}[h]
    \centering
    \includegraphics[width=1\textwidth]{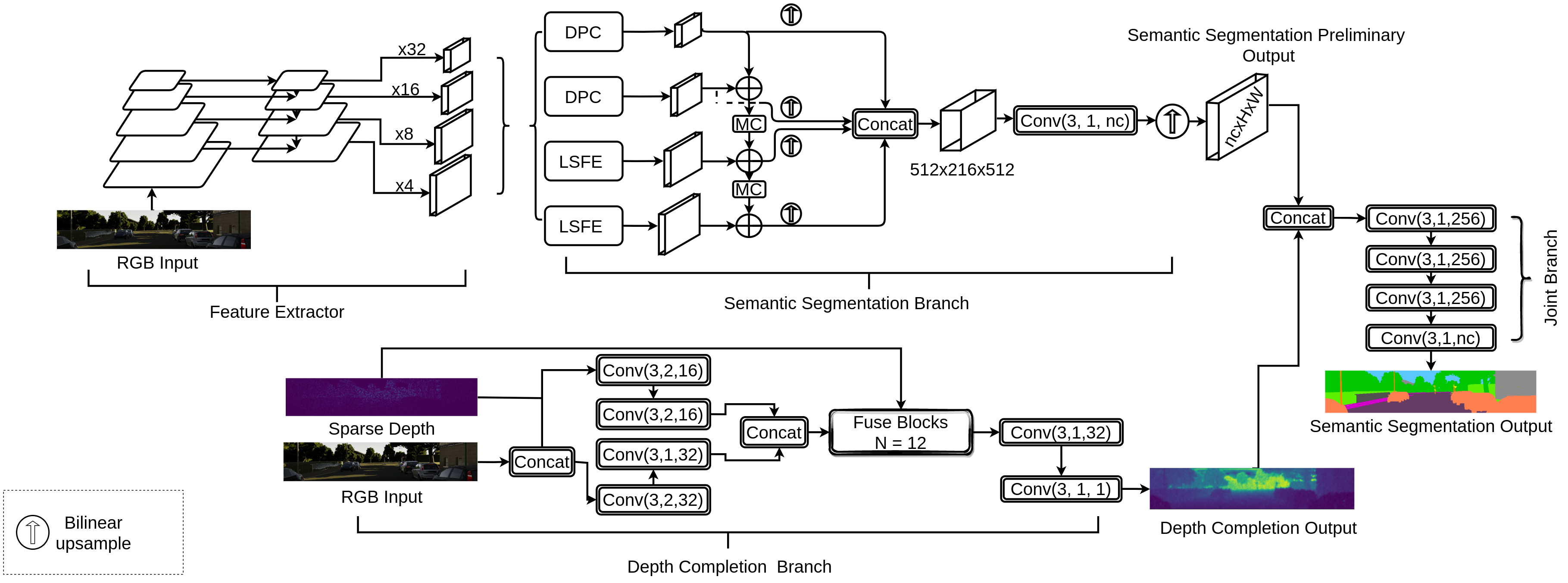}
    \caption{SemSegDepth\_a architecture.}
    \label{fig:app3}
    \end{figure}

\paragraph{SemSeg\_Depth\_a} This model, as shown in Figure \ref{fig:app3}, is based on SemNet\_depth\_gt, where, instead of using the depth completion sparse depth ground truth as input to the joint branch, we predict a dense depth map using DepthNet\_b and pass it then as input to the joint branch. 

\begin{figure}[h]
    \centering
    \includegraphics[width=1\textwidth]{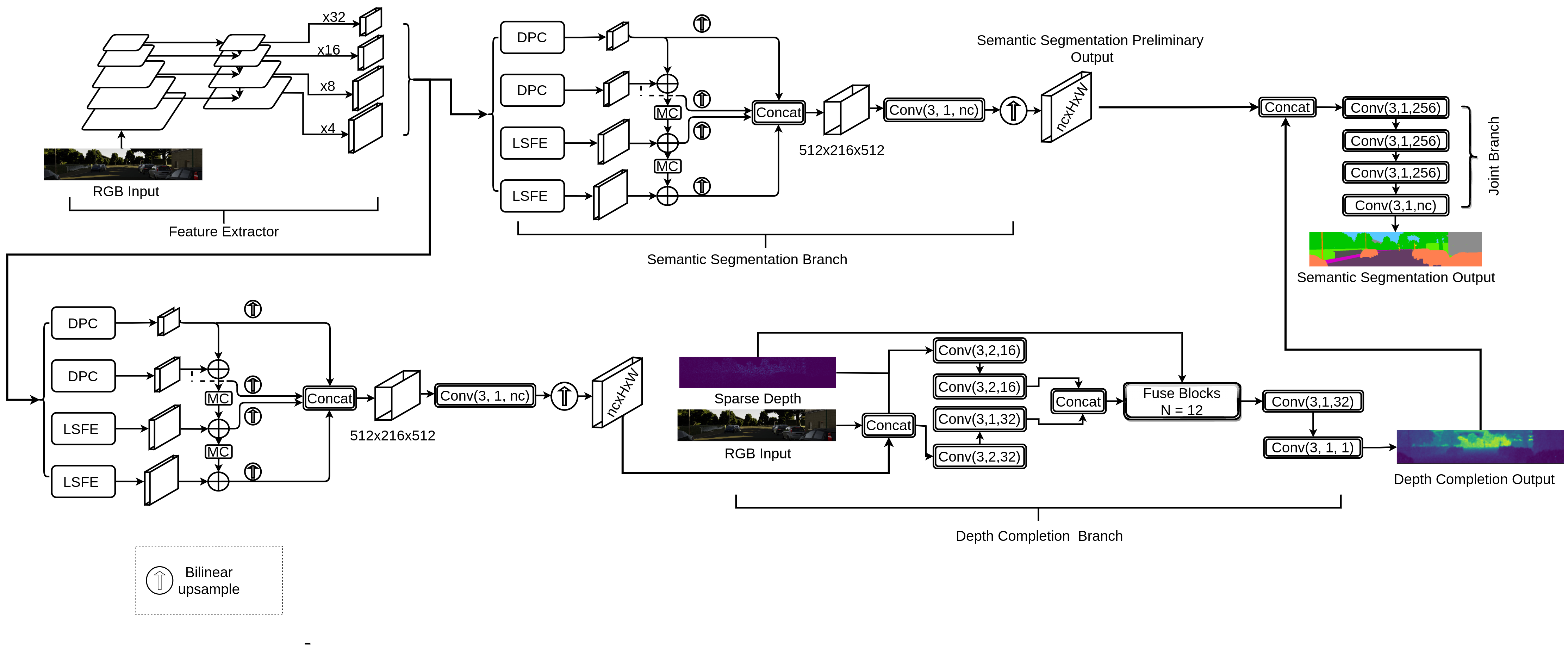}
    \caption{SemSegDepth\_b architecture.}
    \label{fig:app4}
    \end{figure}

\paragraph{SemSeg\_Depth\_b} Similar to SemSegDepth shown in Figure \ref{fig:semseg_depth}. This model combines semantic segmentation and depth completion in a multi-task network. However, different to SemSegDepth, the Semantic Segmentation Preliminary Output is not an input to the depth completion branch. Instead, we use another instance of the semantic segmentation branch to extract semantic features. All in all, this model consists of a feature extractor, two semantic segmentation branches (one of which works as an input to the depth completion branch), a depth completion branch and a joint branch. The architecture of this model is shown in Figure \ref{fig:app4}

\paragraph{SemSeg\_Depth\_c} This model follows the exact same architecture as SemSeg\_Depth\_b. The difference lies in the loss calculation. While in SemSeg\_Depth\_b we only calculate the semantic segmentation and the depth completion loss as in eq. \ref{eq1} and eq. \ref{eq3}, respectively, in {SemSeg\_Depth\_c} we also calculate the joint loss as in eq. \ref{eq4}.

\paragraph{} The quantitative results of all the models used in the ablation studies are shown in Table \ref{tab:abl}.

\begin{table}[h]
\caption{Ablation Experiments.}\label{tab:abl} \centering
\begin{tabular}{|c|c|c|}
  \hline
  Method & mIoU  & RMSE(mm) \\
  \hline
  SemSegNet\_b  & 0.520 & - \\
  \hline
  DepthNet\_b  & - & 580.2 \\
  \hline
  SemNet\_depth\_gt  & 0.542 & - \\
  \hline
  SemNet\_depth\_dense\_gt  & \textbf{0.638} & - \\
  \hline
  DepthNet\_semantic\_gt  & - & 833.7 \\
  \hline
  SemSeg\_Depth\_a  & 0.5421 & 1497.0 \\
  \hline
  SemSeg\_Depth\_b  & 0.5463 & 438.4 \\
  \hline
  SemSeg\_Depth\_c  & 0.5841 & \textbf{429.7} \\
  \hline
  \textbf{SemSegDepth (ours)}  & 0.5932 & 458.2 \\
  \hline
\end{tabular}
\end{table}

It is important to note that neither SemNet\_depth\_gt nor DepthNet\_semantic\_gt are significantly better in terms of their performance, despite having as input the corresponding sparse depth map ground truth and semantic segmentation ground truth respectively. SemNet\_depth\_dense\_gt  outperforms all the other models for semantic segmentation, suggesting that reliable depth data contains information useful for other tasks such as semantic segmentation. However, SemNet\_depth\_dense\_gt relies heavily on a fully dense depth map ground truth as input, which is available in virtual environments only, whereas in real environments such  ground truth is not available, hence sparse depth maps are a far more common.

SemSeg\_Depth\_a shows to perform better on the task of semantic segmentation but much worse performance for depth completion. On the other hand, sharing the backbone weights as in SemSeg\_Depth\_b and SemSeg\_Depth\_c,  demonstrates to be a better performing approach. Furthermore, SemSeg\_Depth\_c outperforms  SemSeg\_Depth\_b by calculating a joint loss as in eq. \ref{eq4}, even when both architectures are exactly the same. 

Finally, in an attempt to share as many weights as possible, our proposed model also shares the weights of the semantic segmentation branch, yielding better performance in the semantic segmentation task. This further highlights the relevance of having two task-specific branches sharing weights in the network as in our proposed model SemSegDepth. Our model learns high-level features containing information  for both tasks intrinsically, which is significantly more accurate as compared to having access to the complementary ground truth.

\section{\uppercase{Conclusions}}
\label{sec:conclusion}

In this paper, we propose an end-to-end  multi-task network for semantic segmentation and depth completion. It combines a modified version of two bench-marking models, more specifically, we used the model proposed by \citet*{chen2020learning} for depth completion and our semantic segmentation branch is based on the semantic segmentation branch of the EfficientPS model proposed by \citet*{Effps}. 

With the proposed model, we successfully provide further evidence that multi-task networks can significantly improve the performance of each individual task by learning features jointly. Our model successfully predicts the fully dense depth map as well as the semantic segmentation image in a scene, given an RGB image and a sparse depth image as inputs to our model. In addition to that, our ablation studies demonstrate quantitatively, that our multi-task network outperforms, by a large margin, equivalent single-task networks.

\section{Acknowledgements}
This work is supported by the Academy of Finland (projects 327910 \& 324346).

\bibliographystyle{apalike}
{\small
\bibliography{main}}

\section*{\uppercase{Appendix}}
All the models introduced in section \ref{ablation} are presented here as appendices.


\onecolumn
    
\end{document}